\documentclass{article}

\usepackage{iclr2022_conference,times}

\usepackage{amsmath,amsfonts,bm}

\def\eqref#1{equation~\ref{#1}}

\def\1{\bm{1}}

\DeclareMathAlphabet{\mathsfit}{\encodingdefault}{\sfdefault}{m}{sl}
\SetMathAlphabet{\mathsfit}{bold}{\encodingdefault}{\sfdefault}{bx}{n}

\newcommand{\E}{\mathbb{E}}

\usepackage[utf8]{inputenc} 
\usepackage[T1]{fontenc}    
\usepackage{hyperref}       
\usepackage{url}            
\usepackage{booktabs}       
\usepackage{amsfonts}       
\usepackage{nicefrac}       
\usepackage{microtype}      

\usepackage{natbib,hyperref}
\hypersetup{ colorlinks, citecolor=green, linkcolor=red, urlcolor=blue}

\usepackage[utf8]{inputenc} 
\usepackage[T1]{fontenc}    
\usepackage{hyperref}       
\usepackage{url}            
\usepackage{booktabs}       
\usepackage{amsfonts}       
\usepackage{nicefrac}       
\usepackage{microtype}      

\usepackage{subfig}
\usepackage{tabularx}
\usepackage{graphicx}

\usepackage{algorithm}      
\usepackage[noend]{algpseudocode} 

\usepackage{pgf}
\usepackage{collcell}
\usepackage{booktabs}

\title{An Efficient Subpopulation-based Membership Inference Attack}

\author{%
  Shahbaz Rezaei\\
  University of California\\
  Davis, CA, USA \\
  \texttt{srezaei@ucdavis.edu} \\
  \And
  Xin Liu\\
  University of California\\
  Davis, CA, USA \\
  \texttt{xinliu@ucdavis.edu} \\
}

\iclrfinalcopy 

\begin{document}

\maketitle
\begin{abstract}
Membership inference attacks allow a malicious entity to predict whether a sample is used during training of a victim model or not. State-of-the-art membership inference attacks have shown to achieve good accuracy which poses a great privacy threat. However, majority of SOTA attacks require training dozens to hundreds of shadow models to accurately infer membership. This huge computation cost raises questions about practicality of these attacks on deep models. In this paper, we introduce a fundamentally different MI attack approach which obviates the need to train hundreds of shadow models. Simply put, we compare the victim model output on the target sample versus the samples from the same subpopulation (i.e., semantically similar samples), instead of comparing it with the output of hundreds of shadow models. The intuition is that the model response should not be significantly different between the target sample and its subpopulation if it was not a training sample. In cases where subpopulation samples are not available to the attacker, we show that training only a single generative model can fulfill the requirement. Hence, we achieve the state-of-the-art membership inference accuracy while significantly reducing the training computation cost.

\end{abstract}

\section{Introduction}
Wide-spread deployment of machine learning models, often trained on private and sensitive data, has raised concerns over privacy of training data \cite{rezaei2021accuracy}. Recently, membership inference (MI) attacks have been introduced to identify if a sample has been used during the training of a victim model or not. 
Existing state-of-the-art MI attacks \citep{sablayrolles2019white, watson2021importance} often rely on training numerous shadow models to estimate the average output of a typical model. Then, any significant deviation from a typical output is an indication of the victim model being trained on the target sample. Despite their decent accuracy, these attacks require training numerous shadow models, which imposes huge computational cost.

In this paper, we propose a fundamentally different MI attack that achieves the same accuracy as SOTA while significantly reducing the shadow training computational overhead. Here, instead of comparing victim model's confidence on the target sample versus the average confidence of typical models, which requires training numerous shadow models, we compare the victim model's confidence on the target sample versus the victim model's confidence on similar samples from the same subpopulation as the target sample. Hence, we obviate the need to train multiple shadow models. In other words, we calculate the expected value of subpopulation confidence rather than the expected value of shadow models' confidence. However, in practice, the attacker may not have access to samples from the same subpopulation. To tackle this issue, we develop a BiGAN-like architecture to train a generator that craft samples from the subpopulation of a given image. In other words, our attack only needs training a single generator model once and then it can be used for even unseen samples.


\section{Attack Overview}
\subsection{Background}
Membership inference attacks aim to distinguish training samples from non-training samples. Training samples are often called \textit{member} samples and non-training samples are called \textit{nonmember} samples. Let $x$, $y$, and $Y_v(.)$ be the target sample, target label, and victim model's output, respectively. Now, let $s(Y,(x,y))$ denote the membership score, where a higher score indicates the higher probability of a sample being member. MI attacks aim to introduce an accurate membership score function.

In the literature, there are two types of MI attacks: 1) with sample calibration, and 2) without sample calibration. The calibration process modifies the membership score of an MI attack such that it takes the target sample's difficulty into account \citep{watson2021importance}. The former attacks do not take the difficulty of the target sample into consideration. They essentially compare the victim model's response on the target sample versus an average response to infer membership (e.g. via shadow models). They use \citep{shokri2017membership, yeom2018privacy} confidence output or loss values of target samples. The intuition is that models are often more confident on their training samples. For example, \cite{yeom2018privacy} uses loss function and \cite{shokri2017membership} uses confidence value as a score function. However, it is shown that this leads to poor attack performance and high false positive mainly because well-generalized models output high confidence on majority of nonmember samples as well \citep{rezaei2020towards}.

The second category of attacks, achieving state-of-the-art MI performance, use some form of sample calibration to distinguish between hard-to-predict member samples from easy-to-predict nonmember samples \citep{watson2021importance}. For instance, \cite{sablayrolles2019white} calibrates the score using the average score both when the target sample is in training data and when it is not, as follows:

\begin{equation}
    s_{Sab}(Y_v,(x,y)) =  s(Y_v,(x,y)) - \frac{\E_{ Y \leftarrow A(x\in D)} [s(Y,(x,y))] + \E_{ Y \leftarrow A(x\notin D)} [s(Y,(x,y))] }{2},
\end{equation}

where $A(c)$ is a randomized training algorithm that samples from the distribution of trained shadow models following a specified condition $c$. $D$ is the shadow model training dataset. The intuition is that if a sample is easy-to-predict, then the calibration term is also large. So, the total membership score is small. However, training overhead of this attack is large particularly if all target samples are not known during shadow training. In that case, each time a new sample is targeted, $\E_{ Y \leftarrow A(x\in D)}$ should be calculated from scratch by training new shadow models. \cite{watson2021importance} tackles this issue by estimating the calibration only on shadow models trained without the target sample. Simply put, their membership score is

\begin{equation}
    s_{Watson}(Y_v,(x,y)) =  s(Y_v,(x,y)) - \E_{ Y \leftarrow A(x\notin D)} [s(Y,(x,y))].
\end{equation}

For the base membership score, $s(.)$, they explored confidence, loss, and gradient norm and showed that loss is slightly outperform others. 

\subsection{Our Subpopulation-based Attack}
\label{sec-our-attack}
All MI attacks discussed above require training numerous shadow models to accurately estimate the expectation in the calibration term and thus are computationally heavy. In contrast, we propose a fundamentally different approach by running the expectation over similar samples rather than similar models. In other words, our attack estimates whether the victim model's loss on the target sample is significantly smaller than the victim models' loss on samples from the same subpopulation whose loss should be similar. We define subpopulation-based score as:

\begin{equation}
    s_{ours}(Y_v,(x,y)) =  s(Y_v,(x,y)) - \E_{ x' \leftarrow G(x)} [s(Y_v,(x',y))],
\end{equation}

where $G(x)$ is a function sampling from the distribution of samples that belong to the same subpopulation as $x$. The benefit of our subpopulation-based approach is that it obviates the need to train numerous shadow models. However, obtaining images from the same subpopulation is not trivial. How to define a subpopulation and to train $G(x)$ is covered in the next section.

\begin{figure*}
\centering
\includegraphics[width = .95\linewidth]{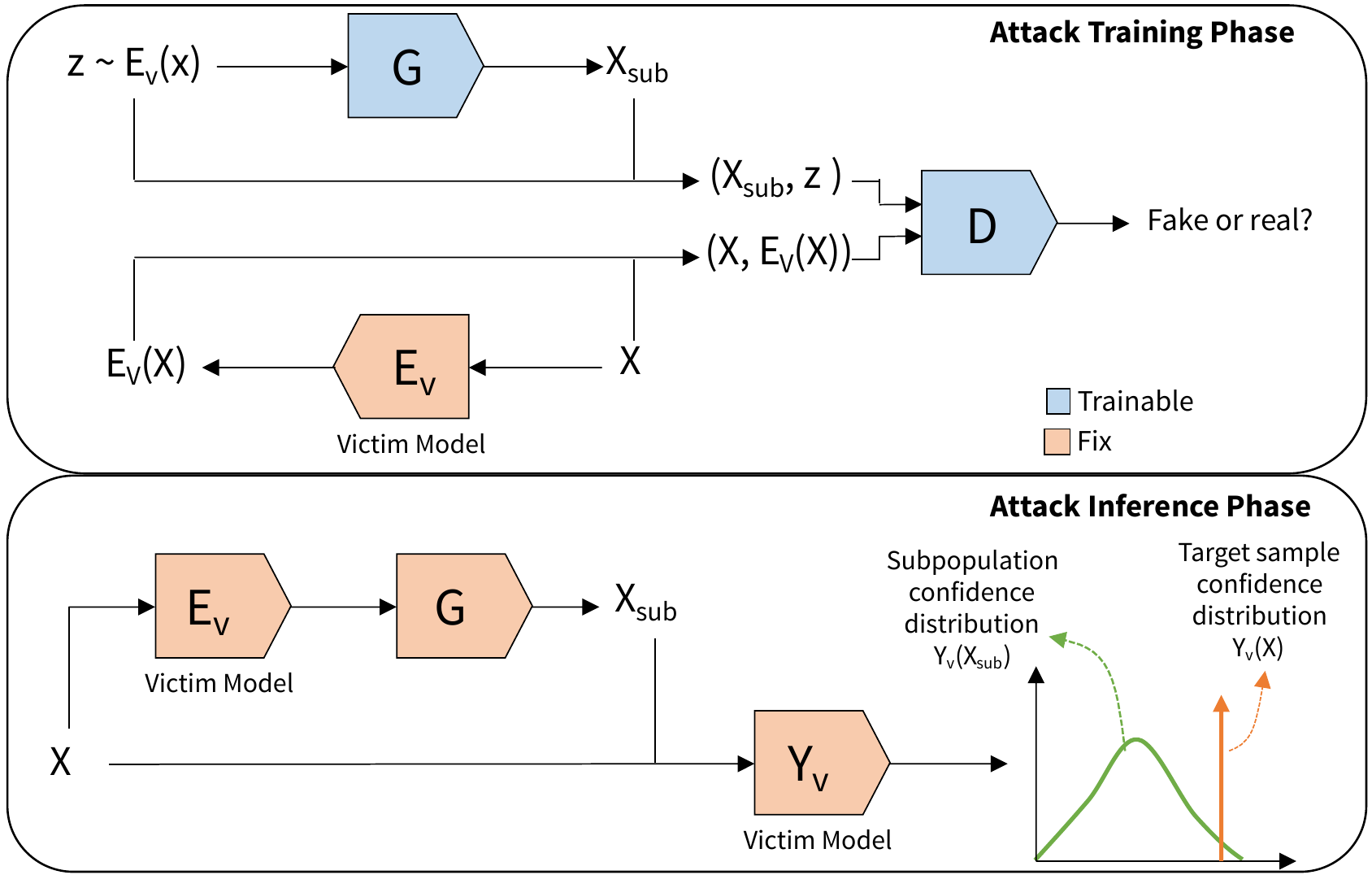}
\caption{Subpopulation-based membership inference attack overview.} 
\label{fig-overview}
\vspace{-.1in}
\end{figure*}

\subsection{Crafting Subpopulations}
\label{sec-attack}
Latent representation of deep learning models have been extensively used as a means of semantic closeness in various applications \citep{vu2020learning, zhang2018visual, yang2019deep}. We define a subpopulation of $x$, $X_{sub}$, such that their samples are close to x in a latent space. Formally, a subpopulation is defined as follows

\begin{equation}
    X_{sub} = \{x' : Dist(E_{v}(x'), E_{v}(x)) < \epsilon  \},
\end{equation}

where $E_v$ is the latent representation of the victim model, and $Dist(.)$ is a distance metric. Here, we consider the output of the last fully connected layer before the softmax as the latent representation. When abundant samples are available to the attacker, she can easily use $X_{sub}$ to launch the subpopulation-based attack as explained in Section \ref{sec-our-attack}. The downside is that the attacker needs to have multiple extra samples for each target sample.

To solve this challenge, we propose a modified version of the BiGAN architecture \citep{donahue2016adversarial} to train a generator model, $G$. The generator learns the mapping from the latent space to the input space which can be used to obtain $X_{sub}$ for MI attacks. However, the original BiGAN architecture cannot be used directly for two reasons: 1) the encoder in BiGAN is trainable while the encoder in our case is the fixed victim model which we cannot change. 2) the original BiGAN forces the encoder to map the input to a uniformly distributed latent space. However, here, the victim model is fixed and there is no guarantee that the latent space follows a uniform or any known distribution. 

Hence, to address these issues, we make the following changes: 
\begin{itemize}
\item We replace the encoder with $E_v$ and block the back-propagation from training $E_v$. 
\item Instead of sampling from a uniform distribution, we obtain the latent representation of all samples in attacker's dataset ($\neq$ victim training dataset) from which we sample as an input for the generator. 
\end{itemize}
See Figure \ref{fig-overview} for an overview illustration of our MI attack.

\section{Experiments}
\label{eval}
We conduct experiments on multiple image classification benchmarks: MNIST \citep{lecun1998gradient}, FMNIST \citep{fmnist}, SVHN \citep{netzer2011reading}, and CIFAR10/CIFAR100 \citep{krizhevsky2009learning}. We divide the train set of these datasets into two parts: victim training dataset and attacker training dataset. The test set is only used for attack evaluation. For MNIST and FMNIST, we choose multi-layer perceptron (MLP) with 4 hidden layers as the victim model. For SHVN, we choose LeNet. For CIFAR10/100 we choose both LeNet and ResNet20. Details of the victim models, generators and discriminators are presented in Appendix \ref{appendix-imp-details}. We train all models using SGD with a learning rate of $0.1$. We reduce the learning rate by a factor of 10 at epoch 50 and 75. The performance of the victim models are shown in Table \ref{tbl-bal-acc-victim}.

We compare our attack with multiple SOTA MI attacks. Unless specified, we follow the same experimental settings to train attack models as suggested in their original paper. For \cite{shokri2017membership}, we train 100 shadow models for all datasets. \cite{yeom2018privacy} attack requires the knowledge about the average training loss to set the threshold which we assume it is known to the attacker for this particular attack. For attack with calibration, including ours, \cite{watson2021importance}, and \cite{sablayrolles2019white}, we use the loss function as the base membership score before calibration. We train 30 shadow models for these two attacks as suggested in \cite{sablayrolles2019white}. We also compare our attack with \cite{jayaraman2020revisiting} because it is similar to our attack in an interesting way: if we define $X_{sub}$ as target samples with random noise, the two attacks would be essentially similar. However, we show that directly adding random noise to input space leads to poor performance. For \cite{jayaraman2020revisiting}, we first use $T=100$ and $\sigma = 0.01$ as suggested in the original paper. Due to the poor performance, we perform random hyper-parameter tuning on $\sigma$, 10 times and we report the best result.

\begin{figure*}
\centering
\centering
\begin{tabular}{cc}
\subfloat[Original images]{\includegraphics[width=0.45\linewidth]{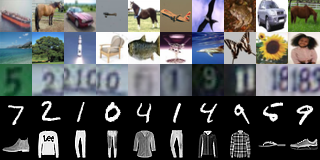}} 
& 
\subfloat[Crafted images from the corresponding subpoplation]{\includegraphics[width=0.45\linewidth]{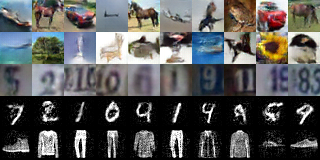}}
\\
\end{tabular}
\caption{Original images (left) and crafted images (right) using BiGAN generator. The datasets/models from top to bottom rows are as follows: CIFAR10/ResNet20, CIFAR100/ResNet20, SVHN/LeNet, MNIST/MLP, and FMNIST/MLP.}
\label{fig-images}
\end{figure*}

For our attack, we consider two scenarios: 1) when a large amount of samples is available to find subpopulation, 2) when training a generator is needed. For the first scenario, we only consider SVHN dataset because it is the only dataset with abundant extra data which is often ignored. We use cosine similarity in latent space to find subpopulations. We find no significant difference when using L2 distance. For the second scenario, we first train a generator in BiGAN-like architecture the details of which is presented in Appendix \ref{appendix-imp-details}. We get the latent representation of each sample using $E_V$. Then, we add small random Gaussian noise, $\epsilon$, with zero mean and standard deviation $\sigma$ proportional to each activation value. In other words, for the latent representation of $x$, denoted as $L = E_V(x)$, noisy latent representation is obtained by $L'_i = L_i + |L_i|\epsilon$. The purpose of the scale factor is to make sure that activations with small values remain small, otherwise it may change the subpopulation or even class of the image. We set $\sigma = 0.05$ for mnist and fmnist, and $0.5$ for other datasets. Finally, we feed the noisy latent representation to the generator to craft a subpopulation for each sample. We craft 30 images per sample. 

Figure \ref{fig-images} illustrates some examples of the target images and their corresponding subpopulation images crafted by our BiGAN generator. Generated images are often similar to the crafted versions with small difference in color, orientation, pattern, background, and texture. Clearly, crafted samples belong to the same subpopulation as the original images. Hence, a victim model output confidence should not be significantly greater than images from the same subpopulation, otherwise it is an indication that the target sample was used during training.

Table \ref{tbl-bal-acc-victim} illustrates the attack performance of all MI attacks. \cite{sablayrolles2019white} outperforms all existing attacks in literature while our attack achieves similar or better performance. Although \cite{watson2021importance} is more efficient, its AUC is lower than \cite{sablayrolles2019white} attack. All previous MI attacks that do not consider sample difficulty (calibration) are substantially worst. Our subpopulation-based MI attack is on a par with \cite{sablayrolles2019white} while obviating the need to train a large number of shadow models. A comparison of computational cost of training multiple shadow model versus a generator model is reported in Appendix \ref{appendix-training-time}. Moreover, our attack achieves a good performance when abundant data is available in which case no model training is needed. It shows that for the membership inference purpose our proposed BiGAN-like architecture achieves the same performance as natural images. 


\begin{table*}
  \caption{Accuracy of victim models}
  \label{tbl-bal-acc-victim}
  \centering
  \begin{tabular}{llllllll}
    \toprule
    \midrule
    Dataset  & MNIST & FMNIST & SVHN  & C-10  & C-100 & C-10  & C-100   \\
    Model  & MLP  & MLP &  LeNet &  LeNet & LeNet & ResNet20 & ResNet20  \\
    Victim train Acc & 100\% & 100\% & 99.89\% & 95.13\% & 95.27\% & 98.51\% & 96.71\%  \\
    Victim test Acc & 97.43\% & 89.59\% & 87.82\% & 57.82\% & 22.37\% & 74.47\% & 33.03\% \\
    \bottomrule
  \end{tabular}
\end{table*}

\begin{table*}
\small
  \caption{AUC of various datasets, target models, and MI attack models}
  \label{tbl-auc}
  \centering
  \begin{tabular}{llllllll}
    \toprule
    \midrule
    Dataset  & MNIST & FMNIST & SVHN  & C-10  & C-100 & C-10  & C-100   \\
    Model  & MLP  & MLP &  LeNet &  LeNet & LeNet & ResNet20 & ResNet20  \\
    \cite{yeom2018privacy} & 51.58\% & 54.84\% & 57.54\% & 77.56\% & 91.98\% & 70.62\% & 93.03\% \\
    \cite{shokri2017membership}   & 51.98\% & 57.69\% & 58.07\% & 75.52\% & 84.72\% & 67.72\% & 87.75\% \\
    \cite{jayaraman2020revisiting}   & 52.20\% & 55.78\% & 56.45\% & 75.01\% & 80.64\% & 68.58\% & 85.97\% \\
    \cite{watson2021importance}     & 54.07\% & 60.52\% & 62.97\% & 80.37\% & 95.47\% & 72.78\% & 93.58\%\\
    \cite{sablayrolles2019white} & \textbf{55.54\%} & 62.55\% & \textbf{63.41\%} & 81.56\% & 96.10\% & 74.84\% & \textbf{95.21\%}\\
    Ours & 54.66\% & \textbf{62.88\%} & 62.05\% & \textbf{81.94\%} & \textbf{96.23}\% & \textbf{75.05\%} & 94.56\% \\
    Ours (Black-box) & 54.12\% & 62.07\% & 61.13\% & 81.24\% & 95.86\% & 74.23\% & 94.24\% \\
    Ours with natural subpopulation & - & - & 61.61\% & - & - & - & -\\
    \bottomrule
  \end{tabular}
\end{table*}

\subsection{Black-box Setting}
To get the subpopulation of a target sample, our attack needs to know the latent representation to generate or find semantically similar samples. This is done by using the last layer before the softmax of the victim model. However, in practice, the victim model's intermediate layers might not be available to the attacker. In this case, the attacker also trains an encoder during the BiGAN training instead of using the victim model. As shown in Table \ref{tbl-auc}, the black-box scenario barely changes the attack performance. 


\section{Conclusion}
In this paper, we propose a fundamentally different approach towards membership inference. Instead of comparing the victims model output versus shadow models' output, we essentially compare the victim model's output on the target sample versus victim model's output on samples from the same subpopulation. This new way of approaching membership inference obviate the need to train dozens to hundreds of shadow models and makes MI attacks more computationally efficient. Moreover, we show that when samples from the same subpopulation is not available, we can train a single generator using BiGAN-like architecture to craft samples of subpopulations. Hence, in the worst case, we only need to train a single generator. Our evaluation results demonstrate that our attack can achieve the state-of-the-art MI attack accuracy with no shadow model training.


\bibliographystyle{iclr2022_conference}
\bibliography{ICLR_2022}

\appendix

\section{Appendix}


\subsection{Full Training Report}
\label{appendix-imp-details}
The MLP model for MNIST and FMNIST consists of 5 hidden layers of size 1024, 512, 256, 128, and 100 with LeakyReLU activation followed by a softmax layer. The last layer before the softmax is used as a latent representation. As the result, the generator has a reverse architecture, starting from an input of dimension 100 followed by 5 layers of size 128, 256, 512, 1024, and 784 (input image size). We use LeakyReLU in the generator as well. For the discriminator, we use an architecture similar to the encoder with a few changes: 1) the input is the concatenation of the input image of size 784 and latent representation of size 100, and 2) the last softmax layer is replaced with a dense layer of 1 neuron as the task is binary classification.

For SVHN, CIFAR10, and CIFAR100 datasets, we train a well-known LeNet model as the victim/encoder model. Here, the internal representation is a vector of size 84. Hence, our generator has input dimension of 84 followed by a dense layer of size 512. Then, the model reshapes it to (1, 1, 512) followed by four convolutional blocks of size 512, 256, 128, and 64. Each convolutional block consists of a 2D Convolutional Transpose layer (filter size of (2, 2) and strides of (2, 2)), LeakyReLU, and 2D Convolutional (filter size of (3, 3)). The number of filters is specified by the block size. However, the number of filters of 2D Convolutional of the last convolutional block is set to 3 to make sure that the output size is (32, 32, 3).

In our experimental evaluations, a shallow discriminator based on LeNet architecture suffers from mode collapse. So, we use a deeper convolutional model consists of four 2D convolutional layers of size 128, 256, 512, and 1024 (filter size of (3, 3) and strides of (2, 2)) followed by LeakyReLU activation after each 2D convolutional layer. Then, we use a flatten layer and concatenate the latent representation here. We find that concatenating latent representation at the middle of the model achieves better convergence than at the beginning of the model. Finally, there is a dense layer of size 64, followed by a LeakyReLU activation and the last dense layer of size 1.

For CIFAR10 and CIFAR100 datasets, we also use ResNet20 as victim/encoder model. However, we use the same generator and discriminator as we use in LeNet case. In our BiGAN training, we find that when a Cosine similarity loss \footnote{The loss is defined in TensorFlow as tf.keras.losses.CosineSimilarity().} is around $20\%$ of the first epoch, the generated images are good enough. For CIFAR10 and CIFAR100, we find that training a GAN model for a few epochs and using the pre-trained generator in the BIGAN architecture prevents mode collapse in BiGAN training. The overall training time is similar since the BiGAN converges faster with a pre-trained generator.

\subsection{Training Cost Comparison}
\label{appendix-training-time}
We use Python 3.6.12 and Tensorflow 2.3, and a server with Intel(R) Xeon(R) Platinum 8168 CPU @ 2.70GHz and NVIDIA GeForce RTX 2080 Ti GPU using Ubuntu 18.04. Table \ref{tbl-training-time} shows the training time required for each attack on minutes using a single GPU instance. Note that we group experiments that has similar training size and model architecture because they essentially have similar training time. As shown in Table \ref{tbl-training-time}, our attack significantly reduces the overall training time overhead. Although training a BiGAN is computationally more expensive than a single discriminator (shadow) model, previous MI attacks require training more than one shadow model. Here, both \cite{watson2021importance} and \cite{sablayrolles2019white} is trained using 30 shadow models, as suggested in \cite{sablayrolles2019white}. Additionally, \cite{sablayrolles2019white} attack requires the average shadow model output of cases where the shadow model is trained with the target sample. Hence, for each new sample to investigate, the MI attack needs to compute $\E_{ Y \leftarrow A(x\in D)} [s(Y,(x,y))]$ from scratch, meaning training 15 new shadow models. Our attack and \cite{watson2021importance} do no require training new models for each new target sample. Moreover, our attack is also more effective when the victim model is deeper, such as in ResNet20. In this case, our generator and discriminator architecture is still the same as the LeNet case. Although it might take longer for the generator model to converge and find the mapping from the latent representation to the input space, it is significantly more efficient than training 30 ResNet20 models.

\begin{table*}
\small
  \caption{Training time comparison of MI attacks in minutes.}
  \label{tbl-training-time}
  \centering
  \begin{tabular}{lllll}
    \toprule
    \midrule
    Dataset  & MNIST/FMNIST & SVHN  & C-10/ C-100 & C-10/C-100   \\
    Model  & MLP &  LeNet &  LeNet & ResNet20  \\
    \midrule
    \multicolumn{5}{l}{When all target samples are known before the attack} \\
    \cite{watson2021importance}     & 46.37 & 86.72 & 56.42 & 281.60 \\
    \cite{sablayrolles2019white} & 46.37 & 86.72 & 56.42 & 281.60 \\
    Ours & 4.97  & 31.26 & 24.73 & 46.36 \\
    Ours without BiGAN & 0 & 0 & 0 & 0 \\
    \midrule
    \multicolumn{5}{l}{Training time per new sample} \\
    \cite{watson2021importance}  & 0 & 0 & 0 & 0 \\
    \cite{sablayrolles2019white} & 23.185 & 43.36 & 28.21 & 140.80 \\
    Ours & 0 & 0 & 0 & 0 \\
    Ours without BiGAN & 0 & 0 & 0 & 0 \\
    \bottomrule
  \end{tabular}
\end{table*}

\end{document}